\documentclass{article}

\usepackage{arxiv}
\makeatletter
\AtBeginDocument{%
  \pagestyle{fancy}%
  \rhead{}%
}

\renewcommand{\keywords}[1]{\par\noindent\textbf{Keywords:} #1}
\makeatother
\usepackage{booktabs}
\usepackage{tabularx}
\usepackage[utf8]{inputenc} 
\usepackage[T1]{fontenc}    
\usepackage{amssymb} 
\usepackage{amsthm} 

\usepackage{url}            
\usepackage{amsfonts}       
\usepackage{microtype}      
\usepackage{graphicx}
\usepackage{amsmath}
\usepackage{array} 
\usepackage{natbib}
\usepackage{lineno}         
\usepackage{longtable}
\usepackage{pdflscape}
\usepackage{orcidlink}

\usepackage{enumitem}
\setlist[itemize]{leftmargin=1.6em}
\setlist[enumerate]{leftmargin=1.8em}

\usepackage{calc}

\usepackage{authblk}

\makeatletter
\renewcommand\AB@affilnote[1]{\textsuperscript{#1}\,}
\makeatother

\usepackage{tikz}
\usetikzlibrary{positioning}
\usetikzlibrary{arrows.meta,positioning,calc,decorations.markings}

\title{Complementarity Without Ground Truth?}

\author[1,2,3]{Andrea Ferrario%
\orcidlink{0000-0001-9968-9474}%
\thanks{Corresponding author. Email: \href{mailto:aferrario@ethz.ch}{\texttt{aferrario@ethz.ch}}}%
\hspace{0.35em}}
\affil[1]{Institute of Biomedical Ethics and History of Medicine, University of Z\"urich, Z\"urich, Switzerland}
\affil[2]{SUPSI, Dalle Molle Institute for Artificial Intelligence (IDSIA), Lugano, Switzerland}
\affil[3]{ETH Z\"urich, Z\"urich, Switzerland}

\author[2,4]{Alessandro Facchini%
\orcidlink{0000-0001-7507-116X}}
\affil[4]{Management in Networked and Digital Societies (MINDS) Department, Kozminski University, Warsaw, Poland}

\author[5,6]{Federico Cabitza%
\orcidlink{0000-0002-4065-3415}}
\affil[5]{Department of Informatics, Systems and Communication, University of Milano-Bicocca, Milano, Italy}
\affil[6]{Digital health and Well-being, Fondazione Bruno Kessler, Trento, Italy}

\begin{document}

\maketitle

\begin{abstract}
Human--AI complementarity captures the idea that a human assisted by an AI system can outperform both the human and the AI acting alone. In prediction tasks, this improvement is assessed against ground truth labels produced through a reference construction procedure and intended to represent the state of affairs to be predicted. Many human--AI interactions, including those where medical AI provides treatment recommendations, however, concern tasks other than predicting what obtains. Their relevant output concerns adequacy relative to a practical goal and depends on several evaluative dimensions. In this work, we examine the ground truth dependence of complementarity and extend it to human--AI interactions where ground truth is not the appropriate reference. We show that tasks beyond prediction can still admit a reference construction procedure that generates adequacy labels. We develop a route for extending complementarity to such settings. It replaces ground truth labels with context-dependent adequacy labels generated through a two-step scoring methodology inspired by multi-criteria decision analysis frameworks in healthcare and it makes the normative choices underlying reference construction explicit. Because the construction of adequacy labels embeds evaluative choices, we draw methodological inspiration from computational reliabilism to identify technical, scientific, and social factors relevant to supporting their construction.
Our approach expands the empirical study of human--AI complementarity beyond prediction tasks, enabling its quantitative evaluation in decision-support settings where ground truth is not the appropriate evaluative reference.
\end{abstract}


\section{Introduction}
\label{section:Introduction}
Since the early 2020s, \emph{complementarity} has become an increasingly investigated performance measure in human--AI interaction (HAI) research \citep{Bansal2021CHI,Bansal2021AAAI,Donahue2022FAccT,cabitza2023rams,Vaccaro2024NHB}. In prediction tasks, a human assisted by an AI system
achieves complementarity when the resulting human--AI team performs better than both the human and the AI alone. This generalises \emph{human augmentation}, where AI improves human performance, and \emph{AI reliance}, where the human adopts AI output \citep{schemmer_appr_reliance,grote2024reliability}.

The conceptual simplicity of complementarity  has, however, exposed a few theoretical and empirical limitations \citep{Vaccaro2024NHB,ferrario2026complementarityextended}. In this work, we focus on one structural assumption: \emph{the dependence of complementarity on ground truth}. 
In prediction tasks, complementarity is evaluated relative to labelled datasets: the predictions of the human, the AI system, and the human--AI team are compared against common reference labels, i.e., the \emph{ground truth labels}, through a loss function or an equivalent label-based performance measure \citep{Donahue2022FAccT,ferrario2026complementarityextended}. Thus, this comparison presupposes a reference label intended to represent which relevant state of affairs obtains and against which the predictions can be scored.

While this setting is appropriate for prediction tasks, HAIs also concern tasks for which this procedure is less natural. Prominently, in the medical field, a clinical AI system may support the \emph{management reasoning} paradigm  \citep{cook2018management,cook2019management}, recommending one treatment among several alternatives, ranking therapeutic options, suggesting escalation or discharge, or supporting the evaluation of competing decision alternatives.
Here, the relevant question is whether an AI-assisted action is adequate within a specific decision context and given several adequacy criteria, such as treatment efficacy, effectiveness, toxicity, and  quality of life. However, none of these dimensions alone determines which treatment should be selected. While these practical judgments still may depend on truths about the world, their evaluation is not exhausted by adequacy with respect to a to-be-predicted state of affairs.

In this work, we address the ground-truth limitation of complementarity through a four-step approach. First, following \citet{hogberg2026position}, we distinguish ground truth from \emph{ground truthing}, the process through which ground truth labels are defined, collected, and validated as references to states of affairs, and we characterise ground truthing as a key component of the reference construction procedure for prediction tasks. Second, using medicine as a source of motivating use cases, and diagnosis and treatment selection as illustrative cases, we generalize this procedure to AI-assisted practical decision-support tasks, exemplified by the management reasoning paradigm in medicine \citep{cook2018management,cook2019management}. We argue that such tasks can also admit labels, but that these encode \emph{adequacy} relative to contextualised evaluative standards rather than correspondence to a state of affairs. Third, we use these adequacy labels to extend complementarity to practical tasks, making the normative commitments involved in defining adequate outputs explicit within the evaluation procedure. Finally, drawing methodological inspiration from computational reliabilism (CR) \citep{duran2025beyond}, we identify factors relevant to supporting adequacy-label construction and examine robustness across admissible reference constructions. The paper makes four contributions:
\begin{enumerate}
\item it characterizes the reference construction procedure underlying complementarity in HAIs; 
\item it generalizes this procedure to practical decision-support tasks through the construction of adequacy labels; 
\item it extends complementarity functionals to these labels while making their evaluative commitments explicit; and 
\item it provides a CR-inspired account of evidential support for reference construction and a robustness criterion for complementarity across admissible reference constructions.
\end{enumerate}

These contributions expand the empirical study of human--AI complementarity beyond prediction tasks, enabling its evaluation in HAIs where ground truth is not the appropriate evaluative reference.

\section{What Is Complementarity?}
\label{section:CY_definition}
In many AI decision-support settings, the relevant question is whether the human assisted by an AI system performs better than either would perform alone \citep{Bansal2021AAAI,Bansal2021CHI}. This motivates 
the definition and measurement of \emph{complementarity}. Formally, a prediction task is a triple
\(\tau:=(X,Y,\ell),\) where $X$ is an input space, $Y$ an output or label space, and
\(
\ell:Y\times Y\rightarrow \mathbb{R}_{\geq 0}
\)
a pointwise loss function. Let
\(D:=\{(x_i,y_i): 1\leq i\leq n\}\)
be a labelled dataset of samples in \(X\times Y\). The \(y_i\) are the \emph{ground truth labels} of the prediction task \(\tau\). For each instance, let
\(\hat y_i^{H},\ \hat y_i^{AI},\ \hat y_i^{HAI}\)
denote the predictions of the human, the AI system, and the \emph{human--AI team}, i.e., the AI-assisted human, respectively. Let $L_H(D)$, $L_{AI}(D)$, and $L_{HAI}(D)$ denote their empirical losses on $D$. The standard choice is to average the loss \(\ell\) across all samples on \(D\). For instance, \(L_H(D):=\frac{1}{n}\sum_{i=1}^n \ell(y_i,\hat{y}^{H}_i)\) and similarly for \(L_{AI}(D)\) and \(L_{HAI}(D)\). Following existing formulations \citep{Donahue2022FAccT,ferrario2026complementarityextended,Hemmer2025EJIS}, complementarity can be represented through the functional 
\begin{align}
\mathcal{C}(D):=\min\{L_H(D),L_{AI}(D)\}-L_{HAI}(D).\label{def:compl_functional}
\end{align}
\emph{The HAI achieves complementarity on $D$ when} \(\mathcal{C}(D)>0\).
Two elements of this definition are relevant here. First, complementarity is defined relative to a labelled dataset. The losses of all three predictors are computed by comparing their outputs with the same ground truth labels $y_i$. Ground truth is therefore part of the evaluative architecture of complementarity, as we will discuss extensively in the forthcoming sections. Second, the formulation of complementarity concerns one human and one AI system interacting under a given protocol. (The protocol is hidden under the term ``human-AI team.'') This setup covers a broad class of current HAIs, including those where a clinician interacts with an AI system, e.g., a radiologist analysing the AI prediction on an MRI image \citep{cabitza2023rams}, but not all of them. Multi-agent complementarity requires additional structure provided, for instance, by tree-based compositions  \citep{ferrario2026tree}. However, we do not address the multi-agent problem here as, from the perspective of complementarity's dependence on a common reference, it is analogous to the human--AI dyad case.

Before proceeding, it is worth noticing that
complementarity extends beyond the reliance paradigm. In reliance settings \citep{schemmer_appr_reliance,grote2024reliability,ferrario2025being}, the final human--AI output is constrained to coincide with one of the two initial predictions:
\(
\hat y_i^{HAI}\in
\{\hat y_i^{H},\hat y_i^{AI}\}.
\)
Complementarity permits interaction outputs outside this set and is applicable to multi-class classification as well as regression problems. 
A further point concerns how complementarity has been formalised and operationalised  across the literature. As Table~\ref{table:complementarity_literature} shows, the dominant status quo for prediction-oriented HAI is reference-relative evaluation on labelled tasks. In the most explicit formal accounts, complementarity is defined through loss, utility, or quantities equivalent to them. In empirical studies, the same idea is often operationalised through accuracy or related diagnostic performance measures computed against common labels. (Broader reviews, however, also include settings---such as creative tasks---that are not naturally captured by a single loss-based prediction formalism \citep{Vaccaro2024NHB}.)
Terminology around complementarity varies as well;  for instance, \citet{Vaccaro2024NHB} introduce \emph{human-AI synergy} as ``where the human-AI group performs better than both the human alone and the AI alone'' \citep[p.~2294]{Vaccaro2024NHB}. More recently, \citet{campagner2026humansmachinestrulycomplement} formalise \emph{Synergy} in probabilistic terms as the condition in which the final combined judgment is more accurate than both the human and the AI considered separately, relative to a common ground truth state. In summary, Table~\ref{table:complementarity_literature} 
suggests two conclusions. First, the formal literature on complementarity in prediction tasks is organised around common evaluative references and, frequently, around explicit loss-based formulations. Second, even when a loss is not written explicitly, many empirical studies still rely on label-based performance quantities, such as accuracy, that can be interpreted as equivalent to simple loss functions. In what follows, we take this prediction-oriented status quo as our starting point.

\section{All About Ground Truth}
\label{section:ground_truth*}

\subsection{Constructing Ground Truth}
\label{subsection:construct_truth}
In science, the expression \emph{ground truth} originated in remote sensing and was subsequently adopted in different disciplines including computer science, psychology,  neuroscience, and even nursing \citep{woodhouse2021ground}. In remote sensing, ground truth is information obtained by direct measurement at ground level that is typically used to verify or calibrate remotely obtained data, e.g., satellite images \citep{woodhouse2021ground}. Nowadays, in a more common formulation, ground truth refers to information about a system that is treated as sufficiently accurate and reliable to calibrate or evaluate a model or procedure \citep{woodhouse2021ground}. In machine learning and, more generally, modern AI, this information is typically encoded through labels against which model or AI system outputs, called \emph{predictions}, are trained or evaluated. In this work, we use \emph{prediction task} in the machine learning sense. Namely, a prediction task is a task in which, given an instance \(x\), an agent produces an output \(\hat y\) intended to represent a task-specified descriptive target concerning what obtains, has obtained, or will obtain, and whose performance can be evaluated against a \emph{reference} \(y\). Unlike its use in the philosophy of science literature, where a prediction is an ``empirical consequence of a theory that had not yet been verified at the time the theory was constructed'' \citep{sep-prediction-accommodation}, prediction in this machine learning sense is temporally neutral. An image classifier may predict whether a camera image shows a wild boar whose presence is simultaneous with, or prior to, the computation of the prediction. A diagnostic system may predict whether a patient currently has a disease. A prognostic system may predict whether the patient will develop sepsis within 24 hours instead. 

To clarify the epistemic, social, and normative structure underlying these predictive tasks, we distinguish four notions: (1) \emph{state of affairs}, (2) \emph{ground truthing}, (3) \emph{ground truth}, and (4) \emph{ground truth label}. We connect them analytically through what we call a \emph{reference construction procedure} for prediction tasks. In developing this procedure, we draw substantially on the position defended by \citet{hogberg2026position}. We summarise their analysis of ground truthing and use it as the basis for our own reference construction procedure and its later generalisation beyond prediction tasks. (Our use of the term \emph{construction}, rather than alternatives such as \emph{making} or \emph{computing}, follows the perspective of \citet{hogberg2026position} on the process of ground truthing.  More on this point in what follows.)  

For the purposes of this work, we use \emph{state of affairs} in the philosophical sense of a possible way things may be \citep{sep-states-of-affairs}. States of affairs may obtain or fail to obtain.\footnote{This distinguishes them from propositions, which are instead true or false.} For instance,
\(\texttt{S}:=\emph{patient \(p\) having melanoma}\) denotes a state of affairs. At the relevant time, $\texttt{S}$ either obtains or does not obtain. The corresponding proposition
``that patient \(p\) has  melanoma'' is true if the relevant state of affairs obtains and false otherwise.\footnote{This is consistent with a standard correspondence-theoretic account according to which a proposition is true when it corresponds to an obtaining state of affairs \citep{sep-states-of-affairs}. Importantly, the state of affairs is not itself a label and is not itself true or false: it is what obtains or fails to obtain.}
In prediction tasks, the state of affairs identifies the prediction target: diagnostic models predict what obtains, while prognostic models predict what will obtain later.



\begin{landscape}
\begin{table}[p]
\centering
\scriptsize
\setlength{\tabcolsep}{3.2pt}
\renewcommand{\arraystretch}{1.10}
\caption{Complementarity and cognate concepts in human--AI interaction, with representative functional forms.}
\label{table:complementarity_literature}

\begin{tabularx}{\linewidth}{
    >{\raggedright\arraybackslash}p{3.5cm}
    >{\raggedright\arraybackslash}p{3.75cm}
    >{\raggedright\arraybackslash}p{2.5cm}
    >{\raggedright\arraybackslash}p{3.75cm}
    >{\raggedright\arraybackslash}p{3.35cm}
    >{\raggedright\arraybackslash}p{1.95cm}
}
\toprule
\textbf{Concept} &
\textbf{Definition} &
\textbf{Measure} &
\textbf{Functional form} &
\textbf{Loss / performance basis} &
\textbf{Source} \\
\midrule

\textbf{Human--machine collaboration} &
Algorithmic output informs a human who retains the final judgment. &
Joint performance &
No functional is introduced; performance is compared across human-only, AI-assisted, and related conditions. &
Empirical performance comparison; typically label-based task performance. &
\citet{vaccaro2019effects}
\\

\textbf{Human--machine complementarity} &
Human and machine capabilities are combined so that team performance can exceed either alone. &
Team utility/loss &
Joint objective for team performance &
Explicit loss / utility-based formulation. &
\citet{wilder2020learning}
\\

\textbf{Complementary team performance} &
The human--AI team outperforms both the human and AI alone. &
CTP &
\(\mathrm{Acc}_{HAI}>\max\{\mathrm{Acc}_{H},\mathrm{Acc}_{AI}\}\). &
Accuracy-based; equivalent to \(0\)-\(1\) loss on labelled tasks. &
\citet{Bansal2021CHI}
\\

\textbf{Complementarity} &
The combined human--algorithm system has lower expected loss than either component alone. &
Expected-loss gain &
\(L_{HAI}<\min\{L_H,L_{AI}\}\). &
Explicit loss-based formulation. &
\citet{Donahue2022FAccT}
\\

\textbf{Human--AI complementarity} &
Joint decision-making produces performance beyond the human- and AI-alone baselines. &
Joint accuracy &
\(\mathrm{Acc}_{HAI}\) compared with \(\mathrm{Acc}_H\) and \(\mathrm{Acc}_{AI}\). &
Accuracy-based; equivalent to \(0\)-\(1\)-loss comparison. &
\citet{inkpen2023advancing}
\\

\textbf{Human--AI collaboration protocols} &
Different protocols determine how human and AI judgments form the final diagnosis. &
Diagnostic performance &
No functional; protocols are compared through team diagnostic performance. &
Accuracy and related diagnostic metrics on labelled medical tasks. &
\citet{cabitza2023rams}
\\


\textbf{Human--AI synergy} &
The human--AI group performs better than both the human alone and the AI alone. &
Meta-analytic synergy effect \citep{Vaccaro2024NHB}; probabilistic accuracy conditions \citep{campagner2026humansmachinestrulycomplement}. &
Meta-analytic comparison with the better stand-alone baseline \citep{Vaccaro2024NHB}; 
\(\Pr(F=X)>\Pr(H=X)\) and \(\Pr(F=X)>\Pr(A=X)\) \citep{campagner2026humansmachinestrulycomplement}. &
\citet{Vaccaro2024NHB}: performance across predictive and creative tasks; 
\citet{campagner2026humansmachinestrulycomplement}: correctness relative to a common ground truth state \(X\). &
\citet{Vaccaro2024NHB}; \citet{campagner2026humansmachinestrulycomplement}
\\

\textbf{Human--AI complementarity} &
Collaboration realises gains arising from differences in information and capabilities. &
CP / CE &
Complementarity potential (CP) and complementarity effect (CE) quantify available and realised gains. &
Accuracy/count-based decision performance; closely related to \(0\)-\(1\)-loss settings. &
\citet{Hemmer2025EJIS}
\\

\textbf{Efficient complementarity} &
Complementarity whose predictive gain exceeds its interaction cost. &
\(\Delta_\tau(D)\) / \(\Delta_\tau^{\mathrm{net}}(D)\) &
\(\Delta_\tau=\min\{L_H,L_{AI}\}-L_{HAI}\), \quad
\(\Delta_\tau^{\mathrm{net}}=\Delta_\tau-\lambda c(D)\). &
Loss-based formulation with additional interaction cost. &
\citet{ferrario2026complementarityextended}
\\
\bottomrule
\end{tabularx}
\end{table}
\end{landscape}


\emph{Ground truthing} is the social process through which a task-relevant reference for a prediction task is constructed \citep{hogberg2026position}. Ground truthing includes specifying the relevant state of affairs, determining what is to be observed, selecting the relevant inputs and output space, collecting evidence, interpreting observations, and validating the resulting reference values \citep{hogberg2026position}. Here, humans, measurement technologies, scientific knowledge, institutional practices, resource constraints, and domain norms contribute to this process. Even direct measurement involves such choices. For instance, a sensor-based ground truth depends on which sensor is used, its resolution and sensitivity, where it is placed, when and how often measurements are taken, how measurements are calibrated, and which observations are retained \citep{hogberg2026position}. These choices may depend on available resources, time constraints, accepted scientific practices, previous successful use of the instrument, and the evidential standards of the domain. Hence, ground truthing is the procedure through which evidence about a relevant state of affairs is turned into a task-relative  reference  \citep{hogberg2026position}. In practice, \citet{hogberg2026position} distinguish four broad families of ground truthing processes: direct measurement of natural phenomena, expert interpretation, crowd consensus, and synthetic data or simulations with ``known truths.'' Across these cases, different combinations of evidence, interpretation, and convention apply and guide the construction of the reference values. In multi-rater settings, a related distinction separates the annotations collected from several raters (the \emph{Diamond Standard}) from the single reference labelling derived from them through a reduction such as majority voting (the \emph{Gold Standard}) \citep{cabitza2020if}. Every reduction discards information, and the quality of the resulting reference can be assessed in terms of its reliability, accuracy, and representativeness \citep{cabitza2020if}.

\emph{Ground truth} is the result of ground truthing, namely, the reference adopted for determining whether predictions about that state of affairs count as correct for a specified prediction task. Ideally, ground truth tracks the obtaining state of affairs sufficiently well for the intended evaluative purpose. This distinction is important because the same phenomenon can support different legitimate ground truths depending on how the prediction task is defined. Consider a fixed camera in a forest that captures a wild boar at foggy dawn. At the level of the scene, the state of affairs ``the wild boar's being present'' obtains. (In fact, the camera captured a wild boar.) Severe fog may nevertheless make the animal unrecognizable in the image. If the task is to reproduce what a human observer can reliably identify from photographs, the case may be excluded or coded as ``\texttt{wild boar:indeterminate}.'' If the task is to detect animals under degraded visual conditions, the image may instead be retained as a positive instance: ``\texttt{wild boar:yes}.'' Therefore, the state of affairs constrains the ground truth, while the ground truthing procedure determines which evidential representation of that state is appropriate for the task. Textual coding provides another example. Sarcasm, deception, or communicative intent cannot generally be established through direct physical measurement alone. Human coders use contextual information, interpretive rules, and annotation manuals to determine the reference against which downstream model outputs are evaluated \citep{braun2006using}. In these cases, the ground truth remains constrained by the phenomenon and available evidence, but its construction requires interpretation through human norms \citep{hogberg2026position}.

Finally, the \emph{ground truth label} is the formal representation of ground truth in the output space \(Y\) of the prediction task \(\tau\). Ground truthing ultimately associates an instance $x_i\in X$ with a reference value
\(y_i\in Y.\) In a binary task, $Y$ may be $\{0,1\}$; in multiclass diagnosis it may consist of a finite collection of diagnostic categories; in regression it may be a numerical value. Once this label has been constructed, the human, the AI system, and the human--AI team can be evaluated by comparing their predictions with the same reference value and compute complementarity functionals.

\subsection{From Prediction to Practical Tasks}
\label{subsection:pred_to_practical_tasks}
The reference construction procedure discussed so far allows computing complementarity in an HAI, by comparing the predictions of the human, the AI system, and the human--AI team against common ground truth labels, given a state of affairs formalised as a problem addressed by ground truthing. However, other HAIs may have different targets. For instance, clinical decision-making includes different types of judgments. Some concern questions such as what is the case, what caused it, or what is likely to happen: diagnosis, prognosis, and the interpretation of clinical tests belong here. However, in more practical tasks, AI outputs may concern which action should be selected rather than predicting whether a state of affairs obtains. We call them \emph{practical decision-support tasks}, namely, those tasks in which the relevant AI system output is a recommendation, ranking, or selection among available actions relative to a specified decision context. Examples beyond medicine include AI-assisted candidate selection, allocation of organisational or public resources, and recommendations among competing operational or policy actions. In this work, we use medicine as the motivating domain and treatment recommendation as the main practical-task example. In medicine, these tasks fall within the \emph{management reasoning} paradigm:

\begin{quote}
Management reasoning involves negotiation of a plan and ongoing monitoring/adjustment of that plan. A diagnosis can usually be established as correct or incorrect, whereas there are typically multiple reasonable management approaches. Patient preferences, clinician attitudes, clinical contexts, and logistical constraints should not influence diagnosis, whereas management nearly always involves prioritisation among such factors. [...] Finally, management is typically more complex than diagnosis. \citep[p.1310]{cook2019management}
\end{quote}

Therefore, management reasoning requires the integration of clinical information, patient preferences, and contextual factors to make decisions about the management of a patient. These decisions include comparison and selection among multiple reasonable and defensible solutions, including treatment selection, further testing, triage, escalation, prevention, and follow-up visits \citep{cook2023management,cook2018management}. While diagnostic, prognostic, and practical tasks can be supported by AI systems embedded in these interactions with different levels of functional autonomy \citep{bitterman2020approaching}, at the current stage, \emph{it is not clear whether the reference construction procedure can be generalised to these practical tasks and, if so, how the resulting reference can support the computation of complementarity functionals such as~\eqref{def:compl_functional}.} This is the problem we will address in the remainder of this work. To better motivate it, we start introducing two examples of medical decision-making to discuss how different medical tasks may relate to the computation of complementarity.

\section{Beyond Predictions: Diagnoses and Treatments in Medical Decision-Making}
\label{section:examples}

\subsection{Predicting: AI-Assisted Diagnosis}
\label{subsection:AI_diagnosis}
Suppose a medical AI system receives information about a patient \(p\) at decision time $t_0$ and outputs probabilities over diagnostic alternatives $D_1, \dots, D_N$, with $N\geq 2$, so that the output space is $Y=\{D_1, \dots, D_N\}$. The task concerns which of the relevant states of affairs \emph{patient \(p\) having \(D_i\) at \(t_0\)}, for $i=1,\dots, N$ obtains. Let $D_2$ be angina \emph{pectoris},
and consider 
\[
\texttt{P1}:= \textit{patient \(p\) having angina \emph{pectoris} at \(t_0\)}.
\]
At $t_0$, \texttt{P1} either obtains or fails to obtain, but which is the case, and more generally which is the correct diagnosis, is still uncertain. Subsequent examinations---for instance laboratory testing, chart review, imaging, or expert deliberation---may provide evidence that $D_2$ was the relevant diagnosis and thus \texttt{P1} obtains. Ground truthing then produces an ex-post reference label \(y=D_2.\) Ideally, the same type of procedure producing this label should also have produced those used to train the diagnostic AI system employed to predict some given patient's 
state.\footnote{The multiclass case does not differ in kind from a binary diagnostic question such as ``Does the patient have condition \(D\)?'', with labels $y\in\{0,1\}$. The output space is larger, but the task remains directed at identifying which state of affairs obtains. 
}
The same logic applies to many prognostic tasks, with one difference, namely that the relevant state of affairs concerns a later interval, so whether it obtains is not yet settled at decision time. Consider 
\[
\texttt{P2}:= \textit{ patient \(p\) developing sepsis within \([t_0,t_{24}]\)}.
\]
A prognostic model outputs a probability that \texttt{P2} obtains. Whether it is so, is assessed after the concerned interval has elapsed, through ground truthing (for instance, by applying established clinical criteria for sepsis to subsequent observation) which yields an ex post reference label $y\in\{0,1\}$.\footnote{Unlike in the diagnostic case, interventions prompted by the prediction may actually affect whether \texttt{P2} obtains, which, to some extent, complicates the use of $y$ as a reference for evaluation. For instance, a high-risk prediction may prompt early treatment that prevents sepsis, so that $y=0$ even if the prediction was correct \citep{pajouheshnia2017accounting}.} 
Diagnostic and prognostic AI therefore provide paradigmatic cases for complementarity \citep{Vaccaro2024NHB,cabitza2020if,cabitza2023rams}. In radiology, for instance, Cabitza et al.\ study human--AI collaboration in the detection of knee lesions from MRI scans and they examine both knee-MRI diagnosis and ECG interpretation under different collaboration protocols \citep{cabitza2023rams}. Related work included in the meta-analysis of \citet{Vaccaro2024NHB} evaluates human--AI performance in tasks such as skin-cancer recognition. Across these cases, the reference construction procedure for prediction tasks can be used and
the resulting label can then be binary or multiclass depending on the task definition. As a result, the human, the AI system, and the human--AI team produce predictions about a common diagnostic or prognostic target, and their outputs can be compared against the same reference label. The computation of the complementarity functional~\eqref{def:compl_functional} follows.

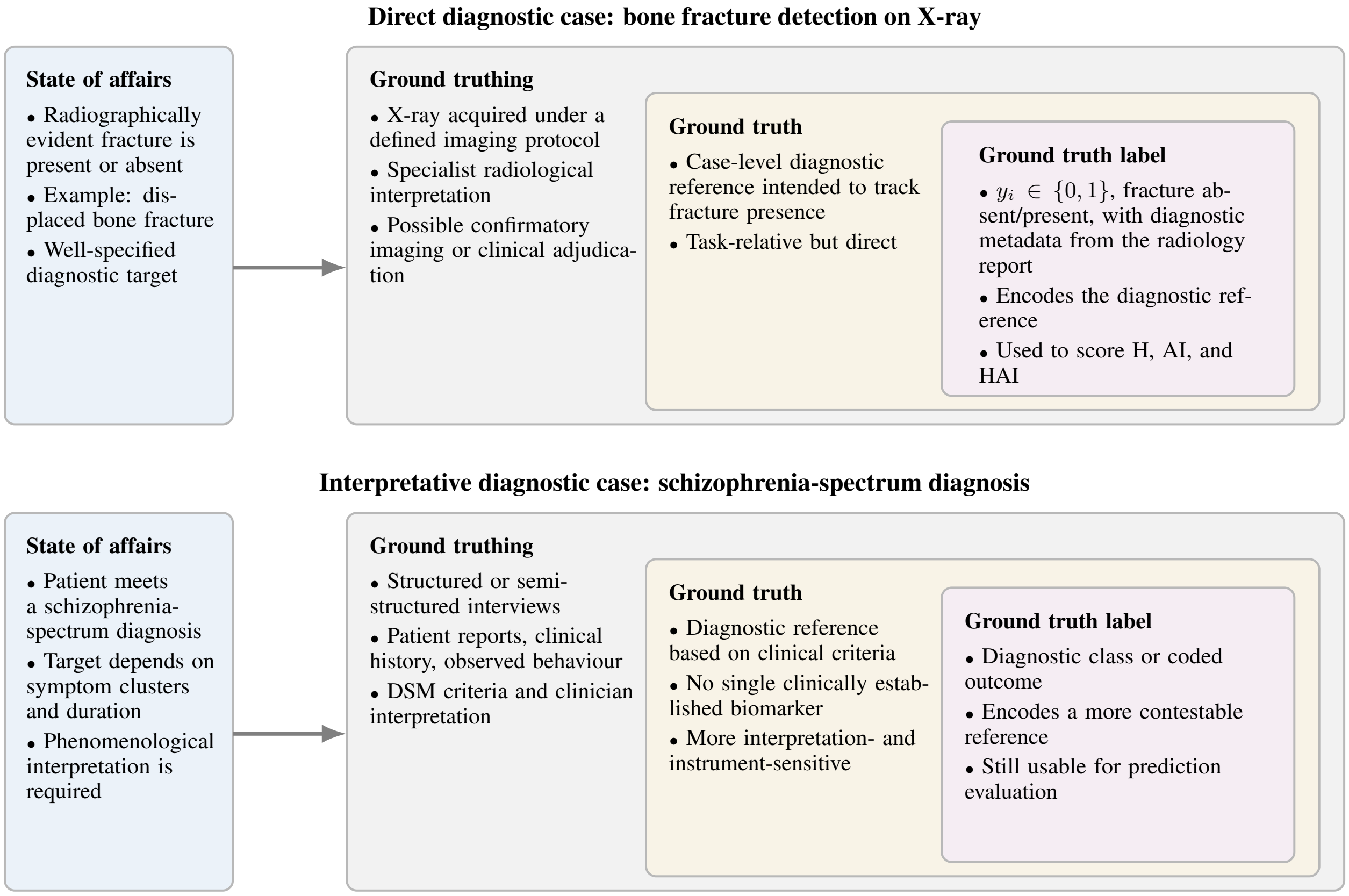
\begin{figure*}[t]
\centering

\definecolor{soaBox}{RGB}{235,242,249}
\definecolor{gtingBox}{RGB}{242,242,242}
\definecolor{gtBox}{RGB}{248,243,231}
\definecolor{gtlabelBox}{RGB}{245,237,243}

\tikzset{
  outerbox/.style={
    draw=black!28,
    rounded corners=4pt,
    line width=0.85pt
  },
  diagtext/.style={
    anchor=north west,
    align=left,
    font=\footnotesize
  },
  diagarrow/.style={
    ->,
    draw=black!50,
    line width=1.55pt,
    >=Latex
  }
}


{\small\textbf{Direct diagnostic case: bone fracture detection on X-ray}}

\vspace{0.45em}

\resizebox{\textwidth}{!}{%
\begin{tikzpicture}[x=1cm,y=1cm]


\draw[
  outerbox,
  fill=soaBox
]
(0,0.9) rectangle (3.2,6.2);

\draw[
  outerbox,
  fill=gtingBox
]
(4.8,0.9) rectangle (18.8,6.2);

\draw[
  outerbox,
  fill=gtBox
]
(9.0,1.1) rectangle (18.45,5.55);

\draw[
  outerbox,
  fill=gtlabelBox
]
(13.15,1.3) rectangle (18.10,5.15);


\node[
  diagtext,
  text width=2.65cm
] at (0.18,5.98) {
\textbf{\small{State of affairs}}\par
\vspace{4pt}
{\tiny$\bullet$} Radiographically evident fracture is present or absent\\[2pt]
{\tiny$\bullet$} Example: displaced bone fracture\\[2pt]
{\tiny$\bullet$} Well-specified diagnostic target
};

\node[
  diagtext,
  text width=3.7cm
] at (5.0,5.98) {
\textbf{\small{Ground truthing}}\par
\vspace{4pt}
{\tiny$\bullet$} X-ray acquired under a defined imaging protocol\\[2pt]
{\tiny$\bullet$} Specialist radiological  interpretation\\[2pt]
{\tiny$\bullet$} Possible confirmatory imaging or clinical adjudication
};

\node[
  diagtext,
  text width=3.6cm
] at (9.2,5.32) {
\textbf{\small{Ground truth}}\par
\vspace{4pt}
{\tiny$\bullet$} Case-level diagnostic reference intended to track fracture presence\\[2pt]
{\tiny$\bullet$} Task-relative but direct
};

\node[
  diagtext,
  text width=3.95cm
] at (13.55,4.92) {
\textbf{\small{Ground truth label}}\par
\vspace{4pt}
{\tiny$\bullet$} \(y_i\in\{0,1\}\),
fracture absent/present,
with diagnostic metadata 
from the radiology report\\[2pt]
{\tiny$\bullet$} Encodes the diagnostic reference \\ [2pt]
{\tiny$\bullet$} Used to score H, AI, and HAI \\
};

\draw[diagarrow] (3.2,3.1) -- (4.8,3.1);

\end{tikzpicture}%
}

\vspace{1.2em}


{\small\textbf{Interpretative diagnostic case: schizophrenia-spectrum diagnosis}}

\vspace{0.45em}

\resizebox{\textwidth}{!}{%
\begin{tikzpicture}[x=1cm,y=1cm]


\draw[
  outerbox,
  fill=soaBox
]
(0,0.9) rectangle (3.2,6.2);

\draw[
  outerbox,
  fill=gtingBox
]
(4.8,0.9) rectangle (18.8,6.2);

\draw[
  outerbox,
  fill=gtBox
]
(9.0,1.1) rectangle (18.45,5.55);

\draw[
  outerbox,
  fill=gtlabelBox
]
(13.15,1.3) rectangle (18.10,5.15);


\node[
  diagtext,
  text width=2.65cm
] at (0.18,5.98) {
\textbf{\small{State of affairs}}\par
\vspace{4pt}
{\tiny$\bullet$} Patient meets a schizophrenia-spectrum diagnosis\\[2pt]
{\tiny$\bullet$} Target depends on symptom clusters and 
duration\\[2pt]
{\tiny$\bullet$} Phenomenological interpretation is  required\\
};

\node[
  diagtext,
  text width=3.7cm
] at (5.0,5.98) {
\textbf{\small{Ground truthing}}\par
\vspace{4pt}
{\tiny$\bullet$} Structured or semi-structured interviews\\[2pt]
{\tiny$\bullet$} Patient reports, clinical history, observed behaviour\\[2pt]
{\tiny$\bullet$} DSM criteria and clinician interpretation
};

\node[
  diagtext,
  text width=3.6cm
] at (9.2,5.32) {
\textbf{\small{Ground truth}}\par
\vspace{4pt}
{\tiny$\bullet$} Diagnostic reference based on clinical  criteria\\[2pt]
{\tiny$\bullet$} No single clinically established biomarker\\[2pt]
{\tiny$\bullet$} More interpretation- and instrument-sensitive
};

\node[
  diagtext,
  text width=3.95cm
] at (13.35,4.92) {
\textbf{\small{Ground truth label}}\par
\vspace{4pt}
{\tiny$\bullet$} Diagnostic class or coded outcome\\[2pt]
{\tiny$\bullet$} Encodes a more contestable reference\\[2pt]
{\tiny$\bullet$} Still usable for prediction evaluation
};

\draw[diagarrow] (3.2,3.1) -- (4.8,3.1);

\end{tikzpicture}%
}

\caption{Two diagnostic instantiations of the reference construction procedure with different levels of complexity. The upper panel illustrates the case of bone fracture detection on X-ray. The lower panel illustrates schizophrenia-spectrum diagnosis, where reference construction relies more strongly on symptom reports, clinical interviews,  and clinician interpretation, while clinically established biomarkers remain limited and diagnostic reliability varies across disorders \citep{yatham2023biomarkers,xie2026test}.}
\label{fig:diagnostic_cases_flow}
\end{figure*}

However, we believe not all diagnostic and prognostic tasks admit equally defensible reference labels. In \texttt{P1} and \texttt{P2}, the relevant criteria, although revisable, are fixed at any given time by established clinical standards. By contrast, psychiatric or syndromic diagnoses, rare diseases with contested or unstable diagnostic criteria, and prognostic targets with uncertain or poorly observable outcomes \citep{yatham2023biomarkers,xie2026test} are more complex. In fact, the adequacy of the ground truth label generated by the reference construction procedure depends, in particular, on how clearly the target state of affairs can be specified and how reliably it can be translated into observable evidence. Thus, even within the prediction task family, the applicability of the reference construction procedure varies, making the conceptualisation and use of ground truth labels dependent on the epistemic and  clinical complexity of the task under consideration.
We show this fact in Figure~\ref{fig:diagnostic_cases_flow}. In the case of a radiographically evident bone fracture, the target state of affairs is comparatively well specified and the reference can be constructed from imaging evidence and expert interpretation. In schizophrenia-spectrum diagnosis, constructing the reference requires more interpretation of symptoms, patient reports, clinical history, and diagnostic criteria. In both cases, however, the reference remains directed at determining which state of affairs obtains.


\subsection{Beyond Predicting: AI-Assisted Treatment Selection}
\label{subsection:AI_treatment}
Consider a medical AI system that recommends one among $N$ possible treatments, \(T_1,\ldots,T_N.\) If complementarity is applied to this case, what would the corresponding ground truth label represent? The problem is that a treatment is not true or false in the same sense in which a diagnostic or prognostic proposition can be true or false. Clinical treatment selection is assessed against several criteria of adequacy, including efficacy, effectiveness, toxicity, adverse effects, quality of life, patient-reported outcomes, treatment burden, and other patient- or institution-specific considerations. The set of admissible options is therefore context-dependent: a treatment may be routinely recommended in one institution but restricted in another because of local guidelines, available expertise, resource constraints, reimbursement rules, or access to specific technologies. These treatment alternatives are  evaluated within a clinical and institutional context.

That said, true or false propositions can be stated about treatments. For instance, the proposition ``that treatment \(T_A\) reduces mortality in population \(P\) by 20\%'' may describe a state of affairs that obtains and may be supported or refuted by evidence. Simply, such a proposition does not by itself answer the question of interest for this AI-assisted decision-making case. This is ``should patient \(X\) receive \(T_A\) (instead of the others)?'' or, in terms of state of affairs,  
\[
\texttt{P3}:= \text{treatment \(T_A\) being the best treatment for patient \(p\)}.
\]

Then, why wouldn't we simply give the proposition describing \texttt{P3} a truth value and record ground truth labels as in the case of \texttt{P1} and \texttt{P2}? \emph{Our claim is that a decision-support task requires a different reference construction procedure than a prediction task}. Let us discuss it in some detail. First, diagnostic AI asks which state of affairs obtains while treatment-recommendation AI supports the selection of an action, in our case the selection of the treatment that is most adequate for a patient---the ``best'' in \texttt{P3}---given relevant criteria. Nevertheless, these more practical tasks also depend on truths about the world, such as patient physiology, comorbidities, and biomarkers. However, for these tasks these truths are not sufficient to determine the practical ranking of recommended treatments. Altogether, patient preferences, risk tolerance, therapeutic aims, clinical norms, institutional constraints, and acceptable trade-offs can affect which alternative is preferable. Thus, while in prediction tasks context and evidence help establish what obtains, in recommendation tasks, some contextual information helps determine how alternatives should be valued or ranked, given a decision goal. Second, even complete descriptive information about the available treatments would not automatically determine a unique best option. Practical adequacy depends on an evaluative relation between the alternatives, the patient, and the clinical context. In simpler diagnostic AI cases, on the contrary, complete descriptive information is typically sufficient to determine a reference label once the relevant diagnostic criteria are fixed. Lastly, treatment evaluation faces an additional empirical problem concerning counterfactual outcomes. After selecting \(T_A\), clinicians may observe the patient's outcome under \(T_A\), but not the outcomes that the same patient would have obtained under \(T_B,\ldots,T_N\). As a result, the evaluation associated to \texttt{P3} is also structurally different from the one of \texttt{P1} and \texttt{P2}: in the former case, only the outcome under the administered treatment is observed, while the outcomes under alternative treatments remain counterfactual. 

 \vspace{1em}

In summary, different AI-assisted tasks are associated with propositions whose content can be more or less directly interpretable and whose corresponding states of affairs can be more or less accessible to evaluation. 
In \texttt{P1} and \texttt{P2}, the propositional content is comparatively well specified and the relevant state of affairs can often be established through observation, testing, or subsequent events.\footnote{ Other predictions may require more interpretation, such as in the case of the proposition  ``that patient \(p\) believes they are a genius.'' Treatment recommendation moves further along this continuum.} They represent propositions that can be tested  by following the reference construction procedure for prediction tasks. In \texttt{P3}, the content of ``best treatment'' is itself evaluative and its assessment is also epistemically different because the outcomes of non-administered treatments remain counterfactual. Nevertheless, the general reference construction procedure for prediction tasks can still be retained. The challenge is to address the changes required at the level of each step of the procedure. We do it in the following section.

\section{The Reference Construction Procedure Beyond Prediction Tasks}
\label{section:ground_truthing_beyond}
We defend the claim that the reference construction procedure for prediction tasks can be generalised to practical decision-support tasks exemplified by AI-assisted treatment recommendations. Perhaps unsurprisingly, the reference construction procedure for these tasks still consists of four components. A \emph{task target} specifies what the interaction is expected to determine. It generalises the specification of the state of affairs for prediction tasks. A \emph{reference construction} specifies how evidence, contextual information, and relevant criteria are used to assess this target. It generalises the ground truthing step discussed so far. The reference construction produces a \emph{task-relative reference}, which determines the standard against which outputs are evaluated; it plays the role of ground truth in prediction tasks. Finally, a \emph{reference label} operationalises this standard in the output space used to compare the human, the AI system, and the human--AI team, generalising the ground truth labels \(y_i\) discussed in the previous sections.

For prediction tasks, this procedure specialises to the one discussed in Section~\ref{subsection:construct_truth}. For AI-assisted treatment recommendation, however, each step takes a different form. The task target may be ``select the most adequate treatment for patient \(p\).'' Then, reference construction requires identifying which treatment alternatives are clinically admissible and which criteria are relevant for comparing them. Here, evidence may include randomised trials, clinical guidelines, patient history, biomarkers, and expected treatment effects. At the same time, patient preferences, local guidelines, reimbursement conditions, and access to particular drugs or procedures may affect the construction of the reference. These elements can also vary across institutions and health systems: a treatment available in one hospital may be unavailable in another, different clinical traditions may tolerate different levels of toxicity, and previous clinical experience may further constrain the set of applicable alternatives. The resulting reference therefore specifies how these elements jointly determine which treatment counts as adequate in the relevant context. Finally, the reference label  records the treatment, ranking, adequacy class, or score produced by this procedure. Its form is discussed in Section~\ref{section:practical_complementarity}.

In summary, \emph{practical tasks can admit reference labels.} The status of these labels depends on the criteria and contextual information used during reference construction. Consequently, \emph{complementarity remains reference-relative, while the reference need not be a ground truth label.} 
The remaining challenges are twofold: (1) to specify how the reference labels generated by the generalised procedure can enter the complementarity functional while preserving its comparative logic, and (2) to assess under which conditions such constructed labels are sufficiently supported as evaluative references.

\section{Reference Construction and Complementarity Functionals for Practical Tasks}
\label{section:practical_complementarity}
We develop an implementation of reference labels for complementarity functionals beyond prediction tasks. It constructs a single adequacy label as the result of a two-step scoring methodology inspired by multi-criteria decision analysis frameworks in healthcare \citep{dolan2010multi,thokala2016multiple}. 
To explain this route, we discuss the case of AI-assisted recommendations for respiratory treatments in an emergency department. The same construction can be adapted to other practical tasks.

\subsection{Constructing Adequacy Labels}
\label{subsection:constructing_adeq_labels}
\begin{figure*}[t]
\centering

\definecolor{targetBox}{RGB}{235,242,249}
\definecolor{constructionBox}{RGB}{242,242,242}
\definecolor{referenceBox}{RGB}{248,243,231}
\definecolor{labelBox}{RGB}{245,237,243}

\tikzset{
  outerbox/.style={
    draw=black!28,
    rounded corners=4pt,
    line width=0.85pt
  },
  diagtext/.style={
    anchor=north west,
    align=left,
    font=\footnotesize
  },
  diagarrow/.style={
    ->,
    draw=black!50,
    line width=1.55pt,
    >=Latex
  }
}

{\small\textbf{Respiratory support recommendation in an emergency department during increased respiratory-infection activity}}

\vspace{0.5em}

\resizebox{\textwidth}{!}{%
\begin{tikzpicture}[x=1cm,y=1cm]


\draw[
  outerbox,
  fill=targetBox
]
(0,-0.5) rectangle (3.6,6);

\draw[
  outerbox,
  fill=constructionBox
]
(5.2,-0.5) rectangle (19.4,6);

\draw[
  outerbox,
  fill=referenceBox
]
(9.7,-0.0) rectangle (19.05,5.8);

\draw[
  outerbox,
  fill=labelBox
]
(14.65,0.3) rectangle (18.75,5.0);


\node[
  diagtext,
  text width=3.0cm
] at (0.18,5.98) {
\textbf{\small{Task target}}\par
\vspace{4pt}
{\tiny$\bullet$} Patient profile \(p_i\)
comprises age,
\(SpO_2\), heart rate,
blood pressure,
comorbidities, etc.\\[4pt]
{\tiny$\bullet$} Common treatment set
\(\mathcal T=\{T_1,T_2,T_3\}\) \\
\(T_1\): conventional oxygen therapy \\
\(T_2\): advanced non-invasive respiratory support\\
\(T_3\): invasive ventilation and ICU escalation
};

\node[
  diagtext,
  text width=3.9cm
] at (5.4,5.98) {
\textbf{\small{Reference construction}}\par
\vspace{4pt}
{\tiny$\bullet$} Decision context \(c\):
ED surge, ICU capacity, staff, equipment, etc.\\[4pt]
{\tiny$\bullet$} Adequacy criteria:
\begin{align*}
&A_1=\text{expected benefit}\\
&A_2=\text{risk of deterioration}\\
&A_3=\text{treatment burden}\\
&A_4=\text{feasibility in context }c
\end{align*}
{\tiny$\bullet$} Established assessments
may inform criteria:
NEWS2, \(\text{SpO}_2\)
};

\node[
  diagtext,
  text width=4.1cm
] at (9.75,5.52) {
\textbf{\small{Task-relative reference}}\par
\vspace{4pt}
{\tiny$\bullet$} Physiological
measurements are first
translated into severity
and deterioration
assessments\\[4pt]
{\tiny$\bullet$} Each treatment \(T_j\)
receives criterion-specific
adequacy values \(v_{ijk}=A_k(T_j,p_i,c)\) \\
{\tiny$\bullet$} These values define
the adequacy profile
\(\mathbf v_{ij}=(v_{ij1},\ldots,v_{ijm})\)\\
{\tiny$\bullet$} The scoring rule
\(s_{ij}=S(\mathbf v_{ij})\)
aggregates the profile
into a common
treatment score
};

\node[
  diagtext,
  text width=3.55cm
] at (14.85,5.00) {
\textbf{\small{Reference label}}\par
\vspace{4pt}
{\tiny$\bullet$} Multiclass adequacy
label \(a_i\in\{1,2,3\}\) computed as in
Equation~\eqref{eq:adequacy_label}\\[4pt]
{\tiny$\bullet$} \(a_i=j\) means that \(T_j\) is the treatment assessed as most adequate for patient \(p_i\)\\[4pt]
{\tiny$\bullet$} Used to compare H, AI, and HAI recommendations
};

\draw[diagarrow] (3.6,3.1) -- (5.2,3.1);

\end{tikzpicture}%
}

\caption{Reference construction for respiratory treatment recommendation in an emergency department. The alternatives follow a typical escalation from conventional oxygen therapy to non-invasive support and, when required, invasive ventilation \citep{smith2019national}.
}
\label{fig:treatment_recommendation_flow}
\end{figure*}
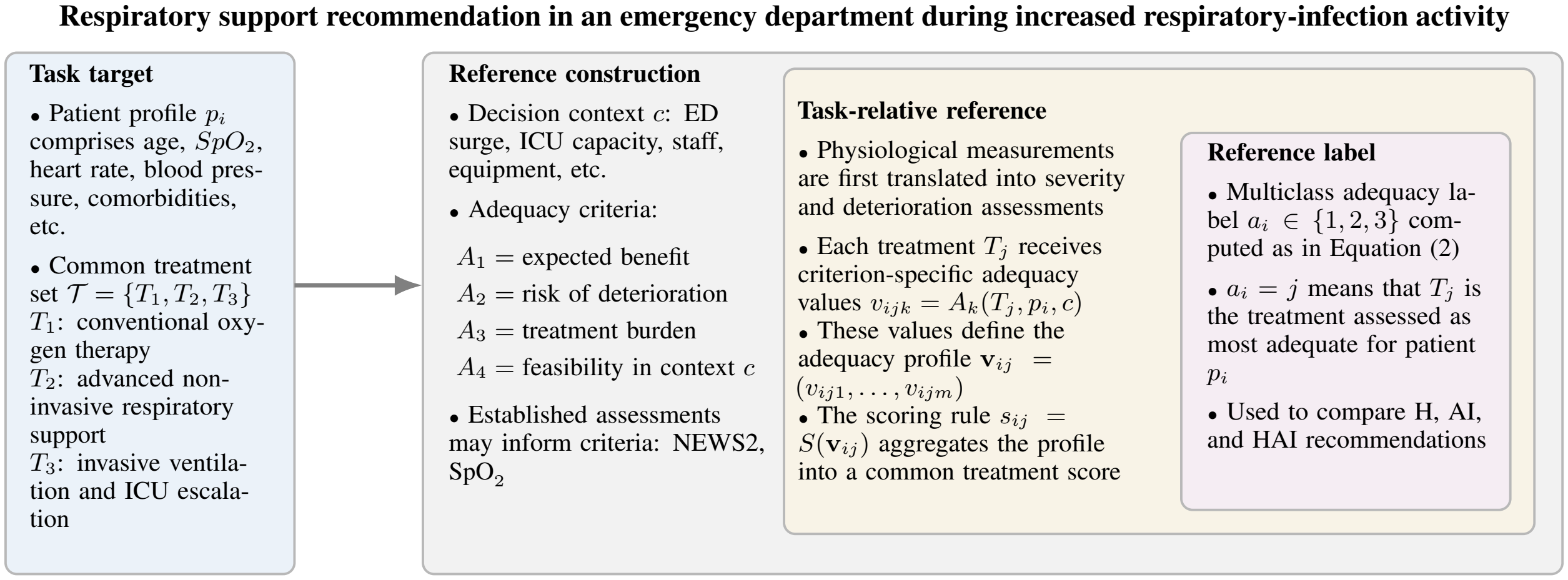

Consider a dataset \(D=\{(x_i,a_i): 1\le i\le n\},\) where \(x_i\) specifies the decision instance and \(a_i\in\mathcal A\) is its \emph{adequacy label}. The goal is to compute these labels using the reference construction procedure for practical tasks introduced in Section~\ref{section:ground_truthing_beyond} and to use them in a complementarity functional. In our AI-assisted treatment recommendation example, we assume that all patients belong to a sufficiently homogeneous clinical population for which the same finite set of treatment alternatives \(\mathcal T=\{T_1,\ldots,T_N\}\) is considered. This set may represent, for instance, the treatment alternatives available within a given ward, disease family, or clinical cohort. Figure~\ref{fig:treatment_recommendation_flow} instantiates this setting for a cohort of patients presenting to an emergency department with respiratory symptoms during a period of increased respiratory-infection activity. The common treatment set contains three levels of respiratory support: conventional oxygen therapy, advanced non-invasive respiratory support, and invasive ventilation with ICU escalation. Reference construction proceeds in two steps. First, each treatment option is assessed \emph{individually} along the relevant dimensions of adequacy, given patient- and context-relative information. Second, these treatment-specific adequacy assessments are combined through a scoring procedure that produces a common ordering of the available treatment alternatives.
This two-step structure is closely related to \emph{Multi-Criteria Decision Analysis} (MCDA) approaches developed for healthcare decision support \citep{dolan2010multi,thokala2016multiple,kremer2021patient}. In these approaches, candidate interventions are first evaluated with respect to several clinically or patient-relevant criteria and the resulting criterion-specific assessments are subsequently combined to obtain an overall score or ranking of the alternatives---see the section on MCDA modeling approaches in \citep{kremer2021patient}. We adapt MCDA by  representing patient-specific information by a vector \(p_i=(p_i^{(1)},\ldots,p_i^{(r)})\in\mathcal P.\)
In the respiratory example, its components include age, oxygen saturation \(\text{SpO}_2\), respiratory rate, heart rate, blood pressure, temperature, and relevant comorbidity indicators. Similarly, the decision context is represented by
\(c=(c^{(1)},\ldots,c^{(q)})\in\mathcal C.\) In Figure~\ref{fig:treatment_recommendation_flow}, the common context includes the emergency-department setting indicator, the current level of respiratory demand, ICU capacity, available staff and equipment levels, and local clinical protocol indicators. (For simplicity, we assume that all decision instances in the dataset \(D\) are evaluated within the same clinical and institutional context.) The precise representation of these components is itself part of the reference construction procedure.

\vspace{1em}

\noindent\textbf{Step I.} In the first step, 
treatment adequacy is assessed through a finite collection of criteria
\(
A_k:\mathcal T\times\mathcal P\times\mathcal C\rightarrow\mathcal V_k,
\ k=1,\ldots,m,\)
where \(A_k(T_{j},p_i,c)\) evaluates treatment \(T_{j}\) with respect to one relevant dimension of adequacy, given patient-relative information and the context, and \(\mathcal V_k\) is the corresponding value space, e.g., \([0,1]\). For each patient \(p_i\), treatment \(T_j\), and adequacy dimension \(k\), define the criterion-specific adequacy value
\(
v_{ijk}
:=
A_k(T_j,p_i,c)
\in\mathcal V_k.
\)
The treatment-specific adequacy profile is then
\(\mathbf v_{ij}:=\bigl(v_{ij1},\ldots,v_{ijm}\bigr)
=\bigl(A_1(T_j,p_i,c),\ldots,A_m(T_j,p_i,c)\bigr)
\in
\mathcal V_1\times\cdots\times\mathcal V_m.
\)
In the respiratory example, these dimensions include expected respiratory benefit, risk of deterioration or organ failure, treatment burden and adverse-event risk, and feasibility in the relevant clinical context. Their assessment may draw on established clinical quantities or scoring systems, including NEWS2 \citep{smith2019national}, and the \(\text{SpO}_2/\text{FiO}_2\) ratio. These instruments inform individual dimensions of the assessment; they do not determine the most adequate treatment by themselves. The criteria, their value spaces, and the rules through which they are combined are specified during reference construction.

\vspace{1em}

\noindent\textbf{Step II.} The second step combines the criterion-specific adequacy values into a common treatment score. Let
\(
S:\mathcal V_1\times\cdots\times\mathcal V_m
\longrightarrow\mathcal S\)
denote an aggregation or scoring procedure, where \(\mathcal S\) is an ordered score space. For treatment \(T_j\) and patient \(p_i\), define
\(s_{ij}:=S(\mathbf v_{ij})=S\!\left(A_1(T_j,p_i,c),\ldots,A_m(T_j,p_i,c)\right).\)
The exact form of \(S\) is task-dependent and may be implemented through a clinical guideline, expert deliberation, an explicit scoring rule, or another auditable methodology. Its purpose is to make the treatment-specific adequacy profiles comparable across the available alternatives. Finally, the adequacy label for patient \(p_i\) is computed by solving (once tie-breaking rules are defined and justified)
\begin{align}
a_i:=\operatorname*{argmax}_{j\in\{1,\ldots,N\}}
s_{ij}=\operatorname*{argmax}_{j\in\{1,\ldots,N\}}
S(\mathbf v_{ij}).
\label{eq:adequacy_label}
\end{align}
Thus, \(a_i\) represents the treatment assessed as \emph{most adequate} under the selected reference construction procedure, given the patient-specific information \(p_i\), the context \(c\), the criterion-specific adequacy profiles \(\mathbf v_{ij}\), and the aggregation rule \(S\).
The respiratory example also clarifies how the adequacy label construction differs from the diagnostic cases represented in Figure~\ref{fig:diagnostic_cases_flow}. Schizophrenia-spectrum diagnosis may require substantial interpretive work before a defensible diagnostic reference can be established. The adequacy-label construction introduces an additional comparative structure. For every treatment \(T_j\), evidence must first be translated into criterion-specific assessments \(A_k(T_j,p_i,c)\); these treatment-specific assessments are then combined through \(S\) to obtain a common ordering of the available actions. The increased complexity therefore concerns the architecture of reference construction: practical treatment selection requires both treatment-specific assessment and an explicit aggregation procedure before the reference label can be produced.


\subsection{Complementarity Functionals With Adequacy Labels}
\label{subsection:compl_adeq_labels}

Let
\(\hat a_i^{H},\ \hat a_i^{AI},\ \hat a_i^{HAI}\)
denote the outputs of the human, AI system, and human--AI team. Given an adequacy loss\footnote{The loss should be chosen consistently with the representation of the treatment recommendation. For hard treatment labels, a $0$--$1$ loss provides the most direct formulation. If agents output probability distributions over treatments, cross-entropy or another proper scoring rule can be used instead.}
\(
\ell_{\text{ad}}:\mathcal{A}\times\mathcal{A}\rightarrow\mathbb{R}_{\geq 0},
\)
define
\begin{align*}
L_Z^{\text{ad}}(D)
:=
\frac{1}{n}
\sum_{i=1}^{n}
\ell_\text{ad}(a_i,\hat a_i^{Z}),
\quad
Z\in\{H,AI,HAI\}.
\end{align*}
Adequacy-relative complementarity is then
\begin{align}
\mathcal{C}^\text{ad}(D)
:=
\min\{L_H^\text{ad}(D),L_{AI}^\text{ad}(D)\}
-
L_{HAI}^\text{ad}(D).
\label{eq:adequacy_complementarity}
\end{align}
\emph{The interaction achieves complementarity relative to the adopted adequacy construction when}
\(
\mathcal{C}^\text{ad}(D)>0.
\)
The downstream comparative structure of complementarity is therefore unchanged. The substantive change lies upstream, in the construction of \(a_i\). Since \(a_i\) depends on the patient-specific information \(p_i\), context \(c\), adequacy criteria \(A_1,\ldots,A_m\), and scoring rule \(S\), every claim based on Equation~\eqref{eq:adequacy_complementarity} is relative to this documented adequacy specification.

\subsection{When are Adequacy Labels Justified?}\label{subsection:label_reliabilist}

Under what conditions should the labels \(a_i\) be treated as adequate references? Transparency alone is insufficient: inspecting the criteria \(A_k\) and aggregation rule \(S\) explains how a label was produced but does not establish that the resulting label is epistemically adequate.\footnote{Here, \emph{transparency} is used in the epistemological sense of understanding a computational process from within. This differs from broader governance uses of the term. For instance, Article~13 of the EU AI Act requires high-risk AI systems to be sufficiently transparent for deployers to interpret their outputs and use them appropriately, together with appropriate information about their characteristics and limitations.}
We address this question by drawing methodological inspiration from computational reliabilism (CR) \citep{duran2025beyond}. According to CR, an agent is justified in accepting the output of a computational process when that process is reliable for the relevant task, where reliability is supported by technical, scientific, and social \emph{reliability indicators} \citep{duran2025beyond}. These indicators are epistemically external to the computational process: although they may shape its specification, validation, and design, its reliability is supported by evidence and practices not exhausted by inspecting its internal computational path \citep{duran2025beyond}.

CR allows computational success to be understood in terms of adequacy for the relevant epistemic purpose \citep{elgin2018true,parker2020model} and has recently been applied to HAIs in predictive tasks \citep{ferrario2026complementarityextended}. The generalised reference construction procedure considered here, however, is not a computational process of the kind to which CR directly applies. It is a broader \emph{socio-technical procedure} that may contain computational subprocesses, e.g., the computation of the scores through the rule \(S\) as in Section~\ref{subsection:constructing_adeq_labels}, but also includes the specification of the decision context, admissible alternatives, evaluative criteria, and aggregation rules through scientific, institutional, and deliberative practices. For instance, clinical evidence, expert practices, patient values, and institutional constraints provide grounds for assessing the quality of such a procedure. However, they also participate directly in defining the criteria \(A_k\), the context \(c\), the admissible alternatives, and the aggregation rule \(S\) through which the adequacy label \(a_i\) is constructed---see Section~\ref{subsection:constructing_adeq_labels}. Thus, some of the epistemic and normative resources that could support an assessment of reliability also partly constitute the procedure whose reliability is under consideration.

For this reason, we do not apply CR directly to the generalised reference construction procedure. Instead, we use its taxonomy of reliability indicators as methodological guidance for assessing whether the procedure is sufficiently well supported for its intended evaluative purpose. Here, technical indicators include agreement between independent assessors on criterion-specific values, stability of \(a_i\) under admissible perturbations of \(S\), validation of criterion-level assessments against observed outcomes, and reliance on established instruments or MCDA practices \citep{cabitza2020if,thokala2016multiple,marsh2016multiple}. In particular, in medical use cases, since criterion-specific values are often elicited from several clinicians, an inter-rater reduction typically precedes the aggregation rule \(S\); case-wise agreement measures such as the degree of weighted concordance can then serve as technical indicators, and the admissible set \(\mathfrak{S}\) may be extended to admissible reductions \citep{cabitza2020if}.
Scientific-practice indicators include grounding the criteria in clinical guidelines, trials, validated instruments, and multidisciplinary expertise. Finally, social indicators include documentation of \(p_i,c,A_k,S\), structured consensus and patient participation, cross-institutional replication, and periodic revision of evidence and clinical standards. The resulting justificatory support is graded, provisional, and local. It is graded because these indicators admit of degrees; provisional because they may be revised in light of new evidence or methodological developments; and local because their relevance depends on the task \(\tau\) and decision context \(c\).

Importantly, this methodological assessment concerns the reference construction procedure and should be distinguished from the downstream application of CR to the HAI  \citep{ferrario2026complementarityextended}. Once a sufficiently supported reference construction has generated the labels \(a_i\), these labels can be used to compute adequacy-relative complementarity through Equation~\eqref{eq:adequacy_complementarity}. Thus,  complementarity occupies an intermediate epistemic role: it presupposes a constructed evaluative reference and, once computed, can function as one reliability indicator of the human--AI process \citep{ferrario2026complementarityextended}. This perspective has two consequences for empirical studies of complementarity. First, any claim that $\mathcal{C}^\text{ad}(D)>0$ should be accompanied by a specification of the task \(\tau\) and of the evidential support for the reference construction from which the adequacy labels were generated. Second, adequacy label construction may admit a set $\mathfrak{S}$ of admissible rules. (For instance, different defensible weighting schemes as scoring rules \(S\).) Let $\mathcal{C}^\text{ad}(D,S)$ denote the complementarity functional computed using labels generated by any $S \in \mathfrak{S}$. An interaction exhibits \emph{robust adequacy-relative complementarity} if 
\[
\min_{S \in \mathfrak{S}} \mathcal{C}^\text{ad}(D,S) > 0.
\]
The admissible set \(\mathfrak{S}\) is itself part of the reference construction and must be independently supported. (We assume that \(\mathfrak{S}\) is finite.) Then, robust complementarity reduces dependence on any single normative choice in \(S\), while its evidential value for HAI reliability remains conditional on the quality of the underlying reference construction \citep{ferrario2026complementarityextended}.

\section{Discussion}
\label{section:discussion}
The construction of adequacy labels for AI-assisted practical tasks extends the scope of complementarity while preserving its comparative structure: the HAI should produce outputs that are better than those of the human or the AI acting alone. The main change concerns the reference relative to which ``better'' is assessed. 
The reference construction procedure for predictive tasks can be adapted to  practical tasks, such as those under the management reasoning paradigm in medicine, but it requires specifying how evidence, patient information, contextual constraints, and evaluative criteria determine which alternatives count as adequate. That said, we do not suggest a sharp separation between prediction and recommendation. For instance, diagnostic and prognostic reasoning may involve substantial interpretation, uncertainty, and contextualisation, as illustrated by the schizophrenia-spectrum case, while some management decisions may be strongly protocol-driven. The distinction relevant to our argument concerns the architecture of the evaluative reference. 
Some prediction tasks admit relatively direct ground truth labels, while others require substantial interpretive work. Practical tasks introduce a stronger evaluative component because the target concerns the adequacy of an action relative to specified goals and criteria. The generalised reference construction procedure discussed in Section~\ref{section:ground_truthing_beyond} provides a common framework for describing these differences.
In particular, the respiratory treatment example from Section~\ref{section:practical_complementarity} should be understood literally as an example, that is, an instantiation of the generalised reference construction procedure we discussed. For the class of practical tasks considered here, the general template requires a specification of the task target and admissible alternatives, relevant instance- and context-specific information, evaluative criteria for comparing the alternatives, and a justified procedure for aggregating these assessments into a task-relative reference. The particular criteria \(A_k\), aggregation rule \(S\), and resulting reference-label space are task-dependent. Other practical tasks may therefore instantiate the same architecture through different evidential sources, criteria, and aggregation procedures. We also note that our approach deliberately preserves the comparative structure and standard loss-based formulation of complementarity while generalising the tasks and reference labels to which it applies. An orthogonal strategy would instead attempt generalising the underlying notion of performance, for instance through utility-, regret-, ranking-, or set-valued measures. We leave such alternative extensions of complementarity for future work.

The proposed reference construction procedure has a direct implication for empirical research on HAI. Once such a reference construction procedure has been specified for a practical task, experimental studies can compare the human, the AI system, and the HAI against the same adequacy reference. This preserves the experimental logic already used in complementarity studies on prediction tasks while extending the class of outputs that can be evaluated quantitatively. 
Currently, empirical evaluation of clinical decision support relies on heterogeneous outcome measures. Studies have evaluated clinical decision-support systems through patient outcomes, process-of-care measures, clinician performance, workload, efficiency, usability, trust, and related behavioural or subjective quantities \citep{bright2012effect,nuutinen2023systematic}. 
These measures answer different questions: patient outcomes may be delayed and influenced by factors beyond the original human--AI decision, while process measures such as guideline adherence or decision time may provide only indirect proxies for decision adequacy.  Similarly, measures of trust, reliance, or acceptance describe important properties of the interaction without directly establishing whether the resulting recommendation was better. Reviews of clinical decision-support systems illustrate this heterogeneity 
\citep{bright2012effect} and more recent work on experimental evaluation of machine-learning-based clinical decision support systems similarly identifies the choice of performance measure and ground truth data as central and interdependent design factors \citep{nuutinen2023systematic}. Thus, the adequacy-relative complementarity considered in this work provides an additional evaluative layer for these studies. 
The resulting functional in Equation~\eqref{eq:adequacy_complementarity} is quantitative and reproducible relative to the adopted reference construction procedure. Its transparency follows from documenting the patient- and context-specific information, adequacy dimensions, and aggregation rule that generate the reference labels. As different institutions or studies may adopt different reference constructions, complementarity results can be compared meaningfully only together with the procedures that generated their references. That said, complementarity should remain one indicator within a broader assessment of the reliability of HAI. As argued by \citet{ferrario2026complementarityextended}, observing complementarity provides evidence about the reliability of a human--AI process for a specified task, while it does not exhaust the epistemic, technical, and socio-technical conditions under which reliance on that process is warranted. Other indicators may concern robustness, calibration, the quality of the underlying evidence, the interaction protocol, distributional change, or the institutional environment in which human and AI judgments are produced \citep{duran2024understanding}. The generalised reference construction procedure developed here contributes to this perspective because it requires the epistemic and normative commitments underlying performance evaluation to be specified explicitly. In practical HAIs, the evidential force of performance measures therefore also depends on making explicit and scrutinising the procedures through which successful performance is defined and measured.

Time to address a few criticisms. First, one may object that  \emph{adequacy labels merely reproduce ground truth under a different name}. However, this would simply overlook the distinction between the role played by a reference in the complementarity functional and the epistemic procedure through which that reference is constructed. Although both kinds of labels can serve as common evaluative references for HAIs, their construction and epistemic interpretation remain different. The point of the generalised reference construction procedure is to make this difference explicit.

Second, one may object that \emph{adequacy-dependent complementarity is arbitrary because it depends on choices made during reference construction}. Yet, reference labels are \emph{constructed} as much as ground truth labels are \citep{cabitza2020if,hogberg2026position}, so constructivity itself does not set adequacy labels apart. What distinguishes the present case is that practical reference construction explicitly incorporates evaluative criteria and their aggregation and what matters in both cases is how well the adopted construction is supported. As argued in Section~\ref{subsection:label_reliabilist}, we use the technical, scientific, and social dimensions identified by CR as methodological guidance for assessing this support, rather than applying CR directly to the generalised reference construction procedure. Complementarity claims should therefore be interpreted relative to the adopted construction and, where several constructions are admissible, tested for robustness across them.

Finally, \emph{aggregating several adequacy dimensions into a single reference label necessarily compresses information and may conceal clinically meaningful trade-offs}. This is a genuine limitation of complementarity, as it depends on the specification of reference labels. The criterion-specific adequacy profile remains available before aggregation, and the scoring rule \(S\) should therefore be documented together with the resulting label. That said, a single-label representation remains useful when the aim is to preserve the empirical comparison among human, AI, and human--AI outputs through a common measure of performance (or loss). While some practical tasks may require partial rankings or multiple admissible actions,  complementarity should be understood as one implementable route for measuring performance in those tasks rather than as the most general representation of the quality of practical judgment.

\section{Conclusions}
\label{section:conclusions}
Complementarity is defined for prediction-oriented HAIs in which human, AI, and team outputs can be compared with common ground truth labels. We have shown that its comparative logic can be extended to practical tasks, such as treatment recommendation, by replacing ground truth labels with adequacy labels generated through an explicit reference construction procedure. Such labels combine evidence, patient- and context-specific information, and evaluative criteria, and their use as evaluative references depends on how well the reference-construction procedure is supported. Making that reference construction explicit, contestable, and empirically assessable extends complementarity beyond prediction while preserving its role as an indicator of human--AI performance. Finally, although complementarity is the focal use case developed here, the same reference construction logic may in principle support other evaluation measures that depend on common reference labels; examining such extensions is left for future work.

\section*{Generative AI Usage Statement}
ChatGPT (GPT-5.6 Sol, OpenAI) and Claude (Opus 5.5, Anthropic) were used during manuscript preparation to assist with language editing and spelling. The former was also used to generate the TikZ code for the figures, perform checks on the references. All AI-assisted outputs were subsequently reviewed, verified, and revised by the authors, who take full responsibility for the final manuscript.

\bibliographystyle{ACM-Reference-Format}
\bibliography{references}

@book{elgin2018true,
  title={{True Enough}},
  author={Elgin, Catherine Z},
  year={2017},
   publisher={MIT Press}
}

@article{parker2020model,
  title={Model evaluation: An adequacy-for-purpose view},
  author={Parker, Wendy S},
  journal={Philosophy of Science},
  volume={87},
  number={3},
  pages={457--477},
  year={2020},
  publisher={Cambridge University Press}
}

@article{marsh2016multiple,
  title={Multiple criteria decision analysis for health care decision making—emerging good practices: Report 2 of the {ISPOR MCDA} {Emerging Good Practices Task Force}},
  author={Marsh, Kevin and IJzerman, Maarten and Thokala, Praveen and Baltussen, Rob and Boysen, Meindert and Kal{\'o}, Zolt{\'a}n and L{\"o}nngren, Thomas and Mussen, Filip and Peacock, Stuart and Watkins, John and others},
  journal={Value in Health},
  volume={19},
  number={2},
  pages={125--137},
  year={2016},
  publisher={Elsevier}
}

@incollection{duran2025beyond,
  title={Beyond transparency: Computational reliabilism as an externalist epistemology of algorithms},
  author={Dur{\'a}n, Juan M},
  booktitle={{Philosophy of Science for Machine Learning: Core Issues and New Perspectives}},
  pages={55--79},
  year={2025},
  publisher={Springer}
}

@article{pajouheshnia2017accounting,
  title={Accounting for treatment use when validating a prognostic model: {A} simulation study},
  author={Pajouheshnia, Romin and Peelen, Linda M and Moons, Karel GM and Reitsma, Johannes B and Groenwold, Rolf HH},
  journal={BMC Medical Research Methodology},
  volume={17},
  number={1},
  pages={103},
  year={2017},
  publisher={Springer}
}

@inproceedings{Bansal2021CHI,
  title={Does the whole exceed its parts? {T}he effect of {AI} explanations on complementary team performance},
  author={Bansal, Gagan and Wu, Tongshuang and Zhou, Joyce and Fok, Raymond and Nushi, Besmira and Kamar, Ece and Ribeiro, Marco Tulio and Weld, Daniel},
  booktitle={{Proceedings of the 2021 CHI Conference on Human Factors in Computing Systems}},
  pages={1--16},
  year={2021}
}

@inproceedings{Bansal2021AAAI,
  title={Is the most accurate {AI} the best teammate? {O}ptimizing {AI} for teamwork},
  author={Bansal, Gagan and Nushi, Besmira and Kamar, Ece and Horvitz, Eric and Weld, Daniel S},
  booktitle={{Proceedings of the AAAI Conference on Artificial Intelligence}},
  volume={35},
  number={13},
  pages={11405--11414},
  year={2021}
}

@inproceedings{Donahue2022FAccT,
  title={Human-algorithm collaboration: Achieving complementarity and avoiding unfairness},
  author={Donahue, Kate and Chouldechova, Alexandra and Kenthapadi, Krishnaram},
  booktitle={{Proceedings of the 2022 ACM Conference on Fairness, Accountability, and Transparency}},
  pages={1639--1656},
  year={2022}
}

@article{Vaccaro2024NHB,
  title={When combinations of humans and AI are useful: A systematic review and meta-analysis},
  author={Vaccaro, Michelle and Almaatouq, Abdullah and Malone, Thomas},
  journal={Nature Human Behaviour},
  volume={8},
  number={12},
  pages={2293--2303},
  year={2024},
  publisher={Nature Publishing Group UK London}
}

@article{Hemmer2025EJIS,
  title={Complementarity in human-AI collaboration: Concept, sources, and evidence},
  author={Hemmer, Patrick and Schemmer, Max and K{\"u}hl, Niklas and V{\"o}ssing, Michael and Satzger, Gerhard},
  journal={European Journal of Information Systems},
  volume={34},
  number={6},
  pages={979--1002},
  year={2025},
  publisher={Taylor \& Francis}
}

@article{vaccaro2019effects,
  title={The effects of mixing machine learning and human judgment},
  author={Vaccaro, Michelle and Waldo, Jim},
  journal={Communications of the ACM},
  volume={62},
  number={11},
  pages={104--110},
  year={2019},
  publisher={ACM New York, NY, USA}
}

@inproceedings{schemmer_appr_reliance,
author = {Schemmer, Max and Kuehl, Niklas and Benz, Carina and Bartos, Andrea and Satzger, Gerhard},
title = {Appropriate reliance on {AI} advice: {C}onceptualization and the effect of explanations},
year = {2023},
isbn = {9798400701061},
publisher = {Association for Computing Machinery},
address = {New York, NY, USA},
booktitle = {{Proceedings of the 28th International Conference on Intelligent User Interfaces}},
pages = {410–422},
numpages = {13},
location = {Sydney, NSW, Australia},
series = {IUI '23}
}

@article{inkpen2023advancing,
  title={Advancing human-{AI} complementarity: {T}he impact of user expertise and algorithmic tuning on joint decision making},
  author={Inkpen, Kori and Chappidi, Shreya and Mallari, Keri and Nushi, Besmira and Ramesh, Divya and Michelucci, Pietro and Mandava, Vani and Vep{\v{r}}ek, Libu{\v{s}}e Hannah and Quinn, Gabrielle},
  journal={ACM Transactions on Computer-Human Interaction},
  volume={30},
  number={5},
  pages={1--29},
  year={2023},
  publisher={ACM New York, NY}
}

@article{duran2024understanding,
  title={From understanding to justifying: Computational reliabilism for AI-based forensic evidence evaluation},
  author={Dur{\'a}n, Juan M and van der Vloed, David and Ruifrok, Arnout and Ypma, Rolf JF},
  journal={Forensic Science International: Synergy},
  volume={9},
  pages={100554},
  year={2024},
  publisher={Elsevier}
}

@article{ferrario2025being,
  title={Being pragmatic about reliance and trust in artificial intelligence},
  author={Ferrario, Andrea},
  journal={{Minds and Machines}},
  volume={36},
  number={1},
  pages={5},
  year={2025},
  publisher={Springer}
}

@article{grote2024reliability,
  title={Reliability in machine learning},
  author={Grote, Thomas and Genin, Konstantin and Sullivan, Emily},
  journal={Philosophy Compass},
  volume={19},
  number={5},
  pages={e12974},
  year={2024},
  publisher={Wiley Online Library}
}

@inproceedings{ferrario2026complementarityextended,
  author    = {Ferrario, Andrea and Facchini, Alessandro and Dur{\'a}n, Juan M.},
  title     = {Epistemology gives a future to complementarity in human-AI interactions},
  booktitle = {Proceedings of the AAAI/ACM Conference on AI, Ethics, and Society},
  volume    = {9},
  number    = {2},
  pages     = {936--947},
  year      = {2026}
}

@inproceedings{hogberg2026position,
  author={H{\"o}gberg, Charlotte and Johnson, Ericka and Wagstaff, Kiri L},
    title={Position: Every ground truth is a human construction, not an objective truth},
  booktitle = {{Proceedings of the 43rd International Conference on Machine Learning}},
  series    = {Proceedings of Machine Learning Research},
  volume    = {306},
  pages = 	 {170544--170556},
  publisher = {PMLR},
  year      = {2026}
}

@article{ferrario2026tree,
  title={Tree-based formalization of multi-agent complementarity in human-{AI} interactions},
  author={Ferrario, Andrea},
  journal={arXiv preprint arXiv:2606.04779},
  year={2026}
}

@article{woodhouse2021ground,
  title={On `ground' truth and why we should abandon the term},
  author={Woodhouse, Iain H},
  journal={Journal of Applied Remote Sensing},
  volume={15},
  number={4},
  pages={041501--041501},
  year={2021},
  publisher={Society of Photo-Optical Instrumentation Engineers}
}

@article{wilder2020learning,
  title={Learning to complement humans},
  author={Wilder, Bryan and Horvitz, Eric and Kamar, Ece},
  journal={arXiv preprint arXiv:2005.00582},
  year={2020}
}

@article{cabitza2023rams,
  title={Rams, hounds and white boxes: Investigating human--{AI} collaboration protocols in medical diagnosis},
  author={Cabitza, Federico and Campagner, Andrea and Ronzio, Luca and Cameli, Matteo and Mandoli, Giulia Elena and Pastore, Maria Concetta and Sconfienza, Luca Maria and Folgado, Duarte and Barandas, Mar{\'\i}lia and Gamboa, Hugo},
  journal={Artificial Intelligence in Medicine},
  volume={138},
  pages={102506},
  year={2023},
  publisher={Elsevier}
}

@InCollection{sep-states-of-affairs,
	author       =	{Textor, Mark},
	title        =	{{States of Affairs}},
	booktitle    =	{{The {Stanford} Encyclopedia of Philosophy}},
	editor       =	{Edward N. Zalta and Uri Nodelman},
	howpublished =	{\url{https://plato.stanford.edu/archives/win2025/entries/states-of-affairs/}},
	year         =	{2025},
	edition      =	{{W}inter 2025},
	publisher    =	{Metaphysics Research Lab, Stanford University}
}

@InCollection{sep-prediction-accommodation,
	author       =	{Barnes, Eric Christian},
	title        =	{{Prediction versus Accommodation}},
	booktitle    =	{{The {Stanford} Encyclopedia of Philosophy}},
	editor       =	{Edward N. Zalta and Uri Nodelman},
	howpublished =	{\url{https://plato.stanford.edu/archives/win2025/entries/prediction-accommodation/}},
	year         =	{2025},
	edition      =	{{W}inter 2025},
	publisher    =	{Metaphysics Research Lab, Stanford University}
}

@article{yatham2023biomarkers,
  title={Biomarkers for clinical use in psychiatry: {W}here are we and will we ever get there?},
  author={Yatham, Lakshmi N},
  journal={World Psychiatry},
  volume={22},
  number={2},
  pages={263--264},
  year={2023}
}

@article{xie2026test,
  title={Test-retest reliability of standardized diagnostic interviews for common adult psychiatric disorders: A systematic review and meta-analysis},
  author={Xie, Weiyi and Nordgaard, Julie and Sheldrick, R Christopher and Ahmad, Juwairiya F and Gomes, Fabiano A and Frey, Benicio N and Duncan, Laura},
  journal={JAMA Network Open},
  volume={9},
  number={5},
  pages={e2615039},
  year={2026},
  publisher={American Medical Association}
}

@article{smith2019national,
  title={{The National Early Warning Score 2 (NEWS2)}},
  author={Smith, Gary B and Redfern, Oliver C and Pimentel, Marco AF and Gerry, Stephen and Collins, Gary S and Malycha, James and Prytherch, David and Schmidt, Paul E and Watkinson, Peter J},
  journal={Clinical Medicine},
  volume={19},
  number={3},
  pages={260},
  year={2019}
}

@article{kremer2021patient,
  title={Patient decision aid based on multi-criteria decision analysis for disease-modifying drugs for multiple sclerosis: Prototype development},
  author={Kremer, Ingrid EH and Jongen, Peter J and Evers, Silvia MAA and Hoogervorst, Erwin LJ and Verhagen, WIM and Hiligsmann, Micka{\"e}l},
  journal={BMC Medical Informatics And Decision Making},
  volume={21},
  number={1},
  pages={123},
  year={2021},
  publisher={Springer}
}

@article{thokala2016multiple,
  title={Multiple criteria decision analysis for health care decision making—an introduction: Report 1 of the {ISPOR MCDA} {E}merging {G}ood {P}ractices {T}ask {F}orce},
  author={Thokala, Praveen and Devlin, Nancy and Marsh, Kevin and Baltussen, Rob and Boysen, Meindert and Kalo, Zoltan and Longrenn, Thomas and Mussen, Filip and Peacock, Stuart and Watkins, John and others},
  journal={Value in Health},
  volume={19},
  number={1},
  pages={1--13},
  year={2016},
  publisher={Elsevier}
}

@article{dolan2010multi,
  title={Multi-criteria clinical decision support: {A} primer on the use of multiple-criteria decision-making methods to promote evidence-based, patient-centered healthcare},
  author={Dolan, James G},
  journal={The Patient: Patient-Centered Outcomes Research},
  volume={3},
  number={4},
  pages={229--248},
  year={2010},
  publisher={Springer}
}

@article{braun2006using,
  title={Using thematic analysis in psychology},
  author={Braun, Virginia and Clarke, Victoria},
  journal={Qualitative Research in Psychology},
  volume={3},
  number={2},
  pages={77--101},
  year={2006},
  publisher={Taylor \& Francis}
}

@article{bitterman2020approaching,
  title={Approaching autonomy in medical artificial intelligence},
  author={Bitterman, Danielle S and Aerts, Hugo JWL and Mak, Raymond H},
  journal={The Lancet Digital Health},
  volume={2},
  number={9},
  pages={e447--e449},
  year={2020},
  publisher={Elsevier}
}

@article{cook2019management,
  title={Management reasoning: {I}mplications for health professions educators and a research agenda},
  author={Cook, David A and Durning, Steven J and Sherbino, Jonathan and Gruppen, Larry D},
  journal={Academic Medicine},
  volume={94},
  number={9},
  pages={1310--1316},
  year={2019},
  publisher={Oxford University Press}
}

@article{cook2023management,
  title={Management reasoning: {E}mpirical determination of key features and a conceptual model},
  author={Cook, David A and Stephenson, Christopher R and Gruppen, Larry D and Durning, Steven J},
  journal={Academic Medicine},
  volume={98},
  number={1},
  pages={80--87},
  year={2023},
  publisher={Oxford University Press}
}

@article{cook2018management,
  title={Management reasoning: {B}eyond the diagnosis},
  author={Cook, David A and Sherbino, Jonathan and Durning, Steven J},
  journal={JAMA},
  volume={319},
  number={22},
  pages={2267--2268},
  year={2018}
}

@article{bright2012effect,
  title={Effect of clinical decision-support systems: a systematic review},
  author={Bright, Tiffani J and Wong, Anthony and Dhurjati, Ravi and Bristow, Erin and Bastian, Lori and Coeytaux, Remy R and Samsa, Gregory and Hasselblad, Vic and Williams, John W and Musty, Michael D and others},
  journal={Annals of Internal Medicine},
  volume={157},
  number={1},
  pages={29--43},
  year={2012},
  publisher={American College of Physicians}
}

@article{nuutinen2023systematic,
  title={Systematic review of the performance evaluation of clinicians with or without the aid of machine learning clinical decision support system},
  author={Nuutinen, Mikko and Leskel{\"a}, Riikka-Leena},
  journal={Health and Technology},
  volume={13},
  number={4},
  pages={557--570},
  year={2023},
  publisher={Springer}
}

@article{campagner2026humansmachinestrulycomplement,
      title={When do humans and machines truly complement each other? Formal conditions and empirical evidence for human-AI synergy}, 
      author={Andrea Campagner and Federico Cabitza and Vincenzo Crupi},
      year={2026},
      journal={arXiv preprint arXiv:2610.27965}
}

@article{cabitza2020if,
  title={As if sand were stone. New concepts and metrics to probe the ground on which to build trustable {AI}},
  author={Cabitza, Federico and Campagner, Andrea and Sconfienza, Luca Maria},
  journal={BMC Medical Informatics and Decision Making},
  volume={20},
  number={1},
  pages={219},
  year={2020},
  publisher={Springer}
}

\end{document}